\documentclass[twocolumn]{article}
\pdfoutput=1

\usepackage{iccv}
\usepackage{times}
\usepackage{epsfig}
\usepackage{graphicx}
\usepackage{amsmath}
\usepackage{amssymb}
\usepackage{booktabs}
\usepackage{diagbox}
\usepackage{nicematrix}
\usepackage{multirow}
\usepackage{makecell}
\usepackage{tabularx}
\usepackage{amsfonts}
\usepackage{amsthm}
\usepackage{marvosym}
\usepackage{authblk}
\usepackage[accsupp]{axessibility}
\usepackage[square, comma, sort&compress, numbers]{natbib}

\usepackage[pagebackref=true,breaklinks=true,letterpaper=true,colorlinks,bookmarks=false]{hyperref}

\usepackage[capitalize]{cleveref}
\crefname{section}{Sec.}{Secs.}
\Crefname{section}{Section}{Sections}
\Crefname{table}{Table}{Tables}

\iccvfinalcopy 

\def\iccvPaperID{9413} 

\ificcvfinal\fi

\begin{document}

\title{Anatomical Invariance Modeling and Semantic Alignment for Self-supervised Learning in 3D Medical Image Analysis}



\author[1,3*]{Yankai Jiang}
\author[1,4*]{Mingze Sun}
\author[1,2]{Heng Guo}
\author[1]{Xiaoyu Bai}
\author[1,2]{Ke Yan} 
\author[1]{Le Lu}
\author[1,2\Letter]{Minfeng Xu}
\affil[1]{DAMO Academy, Alibaba Group}
\affil[2]{Hupan Lab }
\affil[3]{College of Computer Science and Technology, Zhejiang University}
\affil[4]{Tsinghua Shenzhen International Graduate School, Tsinghua-Berkeley Shenzhen Institute, China}
\affil[\Letter]{\tt eric.xmf@alibaba-inc.com}

\renewcommand\Authands{ and }

\maketitle
\renewcommand{\thefootnote}{\fnsymbol{footnote}}
\footnotetext[1]{Equal contribution. This work was done when Yankai Jiang and Mingze Sun were interns at DAMO Academy, Alibaba Group.}

\ificcvfinal\thispagestyle{empty}\fi

\begin{abstract}
Self-supervised learning (SSL) has recently achieved promising performance for 3D medical image analysis tasks. Most current methods follow existing SSL paradigm originally designed for photographic or natural images, which cannot explicitly and thoroughly exploit the intrinsic similar anatomical structures across varying medical images. This may in fact degrade the quality of learned deep representations by maximizing the similarity among features containing spatial misalignment information and different anatomical semantics. In this work, we propose a new self-supervised learning framework, namely \textbf{Alice}, that explicitly fulfills \textbf{A}natomica\textbf{l} \textbf{i}nvariance modeling and semanti\textbf{c} alignm\textbf{e}nt via elaborately combining discriminative and generative objectives. \textbf{Alice} introduces a new contrastive learning strategy which encourages the similarity between views that are diversely mined but with consistent high-level semantics, in order to learn invariant anatomical features. Moreover, we design a conditional anatomical feature alignment module to complement corrupted embeddings with globally matched semantics and inter-patch topology information, conditioned by the distribution of local image content, which permits to create better contrastive pairs. Our extensive quantitative experiments on three 3D medical image analysis tasks demonstrate and validate the performance superiority of \textbf{Alice}, surpassing the previous best SSL counterpart methods and showing
promising ability for united representation learning. Codes are available at https://github.com/alibaba-damo-academy/alice.

\end{abstract}

\section{Introduction}
Since the advent of deep learning, the lack of high-quality annotated data has long been a thorny challenge in medical image analysis, especially for 3D tasks.
Recent research efforts based on self-supervised learning (SSL) shed light on acquiring strong visual representations in an unsupervised manner~\cite{komodakis2018unsupervised,he2020momentum, chen2020big, tian2020makes, chen2020simple, grill2020bootstrap}. 

\begin{figure*}
  \centering
  \includegraphics[width=\linewidth]{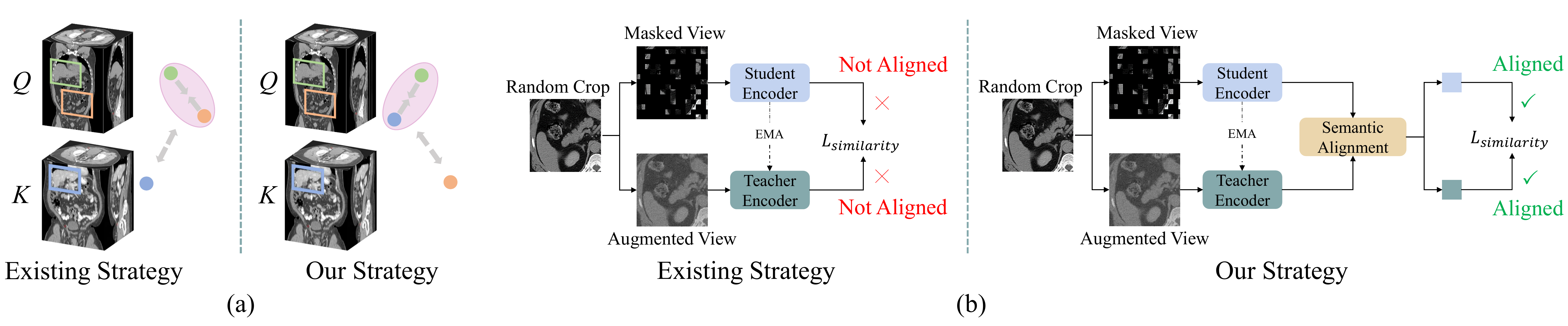}
  \caption{The motivation of our proposed method {\bf Alice}. (a) Existing SSL methods~\cite{caron2021emerging,zhou2021ibot,chen2021empirical, tao2022siamese, huang2022contrastive, tang2022self} simply treat image patch samples which may depict totally different anatomical information from the same CT volume as positive pairs while considering samples sharing the same semantic content class but from another volume as negatives. {\bf Alice} leverages the consistent anatomical structures across different volumes and addresses the false positive and false negative pairs. (b) shows the defect of existing hybrid SSL~\cite{zhou2021ibot, tao2022siamese, huang2022contrastive} methods, which ignore the large semantic gap between masked views and augmented views. Differently, {\bf Alice} performs anatomical information alignment and thus crafts better contrastive pairs.} 
  \label{fig:motivation}
\end{figure*}

Nowadays, contrastive learning (CL)~\cite{caron2021emerging, chen2021empirical, yun2022patch, ranasinghe2022self} and masked image modeling (MIM)~\cite{bao2021beit, xie2022simmim, he2022masked}, together with Vision Transformers (ViTs)~\cite{dosovitskiy2020image,liu2021swin}, have revolutionized the field of SSL in computer vision and medical imaging, which achieve the state-of-the-art (SOTA) performance for a variety of tasks~\cite{caron2021emerging, zhou2021ibot, he2022masked, xie2022unimiss, tang2022self}. There is also a growing trend to combine CL and MIM in a self-distillation way to design more powerful SSL frameworks~\cite{zhou2021ibot, tang2022self, tao2022siamese, huang2022contrastive}.
Despite popularity and success, these methods still follow the self-supervised paradigm designed for specific computer vision scenarios, \eg, ImageNet
(ILSVRC-2012)~\cite{russakovsky2015imagenet}, which can be less suitable or irrational when applied to medical images. 
Now we analyze the drawbacks and irrationalities of existing hybrid SSL approaches, which combine CL with MIM, from the following aspects:

\textbf{(i)} Neglecting the intrinsic similar anatomical structure across varying medical image volumes. 
Commonly-used computed tomography (CT) and magnetic resonance (MR) images render human anatomies with intrinsic structures. 
As shown in~\cref{fig:motivation}a, 
the definition of positive and negative pairs in existing siamese SSL frameworks~\cite{caron2021emerging,zhou2021ibot,tao2022siamese,huang2022contrastive,tang2022self} ignore the semantically consistent anatomical features across different volumes and force an incorrect constraint of instance invariance.
Intuitively, utilizing the intrinsic anatomical structure across different image volumes to model the class-specific invariance can help the learned representations more robust to the size, shape, and texture variances of body parts.

\textbf{(ii)} Lack of anatomical semantic alignment for views extracted or sampled from the same image volume. As shown in~\cref{fig:motivation}b, the widely used siamese architecture in hybrid SSL approaches~\cite{zhou2021ibot, tao2022siamese, huang2022contrastive, chen2022sdae} maximize the similarity between the representations of masked view and intact view. However, a random crop of a body part may contain different organs and human tissues (some of which are small in scale). A large masking ratio would already erase these contents and make the masked  view quite distinct from the intact one.
Thus, maximizing the similarity between views which incorporate totally different semantics can be harmful to the learned representations. 
To learn good representations for downstream tasks, SSL methods for medical images should align the anatomical features of the views which form a positive pair.

Driven by the aforementioned limitations, we present a simple, effective, and dedicated self-supervised learning framework for 3D medical segmentation tasks, {\bf Alice}, by explicitly fulfilling \textbf{A}natomica\textbf{l} \textbf{i}nvariance modeling and semanti\textbf{c} alignm\textbf{e}nt through elaborately combined contrastive learning and MIM. From~\cref{fig:motivation}, {\bf Alice} leverages the structural similarity across different volumetric images 
to explicitly learn universal consistent features from intrinsic body structures and model the anatomical invariance which is robust to the size, shape, intensity, and texture diversity of body parts caused by inter-subject variation, organ deformation, and pathological changes. 


Moreover, we 
design a conditional anatomical feature alignment module which complements masked views with the globally matched anatomical semantics and inter-patch topology to craft better contrastive pairs, enforcing that the positive pairs encoded to semantically consistent feature representations. This process explicitly realizes anatomical semantic alignment to further strengthen the representation with spatial sensitivity and semantic discriminability.

To adequately validate the effectiveness of {\bf Alice}, we employ 3D medical image segmentation and classification as downstream tasks. We fine-tune these widely used ViT-based medical image segmentation frameworks following~\cite{zhou2021nnformer, hatamizadeh2022unetr, tang2022self} with our pre-trained weights on two publicly available benchmarks: Fast and Low-resource semi-supervised Abdominal oRgan sEgmentation in CT (FLARE 2022)\footnote{https://flare22.grand-challenge.org/}, and Beyond the Cranial Vault (BTCV) ~\cite{landman2015miccai}. Our method achieves the current SOTA results, with $86.87\%$ Dice on FLARE 2022 and $86.76\%$ Dice on BTCV, surpassing previous best results by $2.22\%$ and $1.77\%$ respectively. We also evaluate transfer learning on a public 
COVID-19 classification benchmark~\cite{morozov2020mosmeddata}. {\bf Alice} outperforms state-of-the-art counterpart methods by $2.52\%$ in AUC.

Our main contributions can be summarized as:
\begin{itemize}
\item 
We investigate the irrationalities of commonly used siamese SSL frameworks applied to medical images. We propose {\bf Alice} that is customized to leverage the anatomical similarity across volumetric medical images to model class-specific invariance. 
\item In {\bf Alice}, a conditional anatomical semantic alignment module is proposed to match the most related high-level semantics between the crafted contrastive views. 
\item 
{\bf Alice} consistently outperforms popular SSL methods on three public downstream benchmarks, showing its effectiveness and generality.
\end{itemize}

\label{sec:intro}
\section{Related Work}
\noindent \textbf{Self-supervised Learning.}
Contrastive learning is the most popular method in self-supervised learning, which achieves remarkable performance on downstream tasks in computer vision~\cite{he2020momentum, chen2020big, grill2020bootstrap, chen2020simple, chen2020improved, chen2021empirical, caron2021emerging}.
The essence of contrastive learning aims to learn view-invariant representations by maximizing the similarity between the features extracted from different crops of the same image. 
More recently, some works explore how to learn representations by combining contrastive learning and MIM~\cite{tao2022siamese,zhou2021ibot,huang2022contrastive}. As pointed out in~\cite{peng2022crafting}, these methods adopt random sampling to make different crops of the same image, which overlooks the semantic information and may generate views truly contain different image contents. A few newer investigations~\cite{selvaraju2021casting, peng2022crafting, zhang2022leverage} attempt to leverage semantic guided information to crop semantically consistent views. However, 
they ignore the potential positive pairs in other images, restraining the diversity of learned representations. A key difference between these methods and {\bf Alice} is that we mine diversified contrastive views with semantic similar contexts across varying images to encode the intrinsic structure of consistent anatomy information and facilitate the class-specific invariance.

\noindent \textbf{Masked Image Modeling} (MIM) accepts input image corrupted by masking and predicts the target of the masked content, which has been actively studied recently in  self-supervised learning. Existing work mainly differ in their regression objectives~\cite{van2017neural,bao2021beit,dong2021peco,wei2022masked,he2022masked,xie2022simmim} or masking strategies~\cite{kakogeorgiou2022hide, li2022semmae, shi2022adversarial}. 
In this work, {\bf Alice} takes one step further by exploiting cooperations between contrastive learning and MIM to learn effective representations with both strong instance discriminability and local detail sensitive perceptibility, from varying or different image views.

\noindent \textbf{Self-supervised Learning in Medical Imaging.}
Many works apply tailored contrastive SSL methods to medical image problems ~\cite{komodakis2018unsupervised, chaitanya2020contrastive, taleb20203d, xie2022unimiss} with reasonable success. Similar to MIM, image restoration is also commonly used as pre-text task to memorize spatial context from medical images. Typical attempts include inpainting tasks ~\cite{chen2019self,azizi2021big,zhou2021models}, Rubik’s cube problem~\cite{zhu2020rubik} and diverse context reconstruction~\cite{zhou2021preservational}.
Most recently, DiRA~\cite{haghighi2022dira} employs discriminative~\cite{chen2020simple}, restorative~\cite{chen2020generative}, and adversarial learning~\cite{donahue2019large} objectives simultaneously in a unified SSL framework for medical image analysis. Swin UNETR~\cite{tang2022self} trains a transformer-based encoder with combination of different pre-text tasks for 3D medical image segmentation. These efforts constitute important steps toward better SSL methods for medical image analysis.
Nevertheless, {\bf Alice} distinguishes itself by having two key new developments: (1) explicitly leveraging the inherent anatomical consistency between different image volumes to encode the class-specific invariance; and (2) fulfilling anatomical semantic alignment to craft better contrastive pairs.

\begin{figure*}
  \centering
  \includegraphics[width=\linewidth]{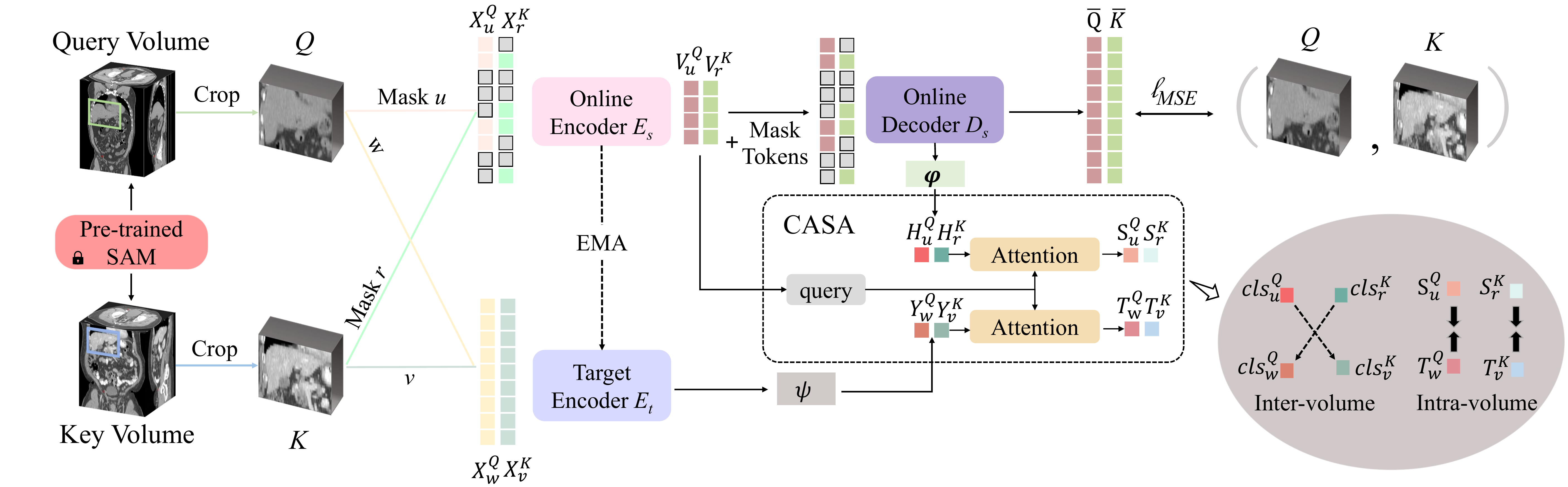}
  \caption{Overall pipeline. {\bf Alice} contains three components: the online encoder, target encoder and online decoder. It first acquires two registered image patch crops from different CT volumes, using SAM \cite{yan2022sam}. Then four different augmented views from crops are fed into the online and target encoders respectively. The online encoder randomly masks a fraction of the image patches and operates on the remaining visible image content. The target encoder operates on the whole view. The online decoder learns to reconstruct the input volume. We propose a conditional anatomical semantic alignment (CASA) module to craft better contrastive pairs. After the pre-training, only the online encoder is kept for downstream segmentation tasks.}
  \label{fig:framework}
\end{figure*}
\section{Method}
The overall framework of our method is illustrated in~\cref{fig:framework}. Our {\bf Alice} model consists of two branches performing masked image modeling and contrastive learning, respectively. 
We first mine diverse yet semantically consistent crops from varying image volumes. 
A query volume and a key volume are randomly picked. Then we adopt a pre-trained SAM~\cite{yan2022sam} model, which performs self-supervised universal landmark detection to locate the same body part in different volumetric medical images and produce two crops, denoted as $Q$ and $K$, that depict two sub-volumes from the same body part. After that, we utilize two different data augmentations, $u$ and $w$, to generate two views of $Q$. We denote them as $X_u^{Q}$ and $X_w^{Q}$. Likewise, we also utilize two different data augmentations, $r$ and $v$, to generate two views of $K$, denoted as $X_r^{K}$ and $X_v^{K}$. Here, $u$ and $r$ are random masking, which is similar to the ``random sampling'' adopted in MAE~\cite{he2022masked}. $w$ and $v$ are two different data augmentation operations. In the following, we elaborate on each component of {\bf Alice} in details and describe how we 
optimize the relationships between and among these four views.
\subsection{Network Architecture Components}
\noindent \textbf{Online Encoder.}
The online encoder $E_s$ takes masked view pairs ($X_u^{Q}$ and $X_r^{K}$) as inputs. Following MAE~\cite{he2022masked}, we only feed the visible, unmasked patches to the online encoder $E_s$. 
$E_s$ embeds visible tokens added with the positional embeddings and produces the output features ($V_u^Q$ and $V_r^K$) through a sequence of transformer blocks.
In this paper, we mainly consider two vision Transformer architectures (ViT~\cite{dosovitskiy2020image} and Swin Transformer~\cite{liu2021swin}) as the online encoder options. We adopt the strategy in~\cite{huang2022green} to allow the Swin Transformer to discard masked patches and operate only on the visible ones. 
After pre-training, only the online encoder $E_s$ is used for extracting image representations in downstream tasks.

\noindent \textbf{Online Decoder.}
In addition to the features of visible patches $V_u^Q$ and $V_r^K$, the online decoder receives mask tokens  
as inputs. We add positional embeddings to all tokens. Then the online decoder learns to reconstruct the pixel of the masked patches. Following MAE~\cite{he2022masked}, we use the normalized pixel values as targets in the reconstruction task. Our loss function $\ell_r$ computes the mean squared error (MSE) on masked patches between the decoder predictions ($\Bar{Q}$, $\Bar{K}$) and the original input volume crops $(Q, K)$:
\begin{equation}
  \ell_r = \frac{1}{n_m^Q}\sum[\Theta(\Bar{Q}-Q)]^2 + \frac{1}{n_m^K}\sum[\Theta(\Bar{K}-K)]^2,
  \label{eq:4}
\end{equation}
where $\Theta$ is an indicator to select the prediction corresponding to masked tokens, $n_m^Q$ and $n_m^K$ are the number of masked patches in $Q$ and $K$, respectively. This MIM objective helps the learned representations encode local context and patient-specific information of input volumes. The online decoder is set to be stacked transformer blocks.

\noindent \textbf{Target Encoder.}
Following existing siamese frameworks~\cite{zhou2021ibot, tao2022siamese, huang2022contrastive}, we introduce a target encoder to generate contrastive supervision for the online encoder to further strengthen the representation learned by MIM with semantic discriminability. 
The target encoder shares the same architecture as the online encoder. We update parameters of the target encoder using an exponential moving average (EMA) of the online encoder weights.
The target encoder takes two unmasked augmented views $X_w^Q$ and $X_v^K$ as inputs and embeds them into high dimensional feature representations which reserve the semantic integrity. 

\subsection{Anatomical Invariance Modeling}
\label{AIM}
A prominent difference between previous methods and {\bf Alice} is that we mine diversified views from different volumes to learn anatomical features that are universal across similar body parts. While existing methods~\cite{zhou2021ibot, tang2022self, huang2022contrastive} only operate on the same volume, we propose to explicitly model the anatomical invariance by maximizing the similarity between embeddings of views from $Q$ and $K$. 

As shown in~\cref{fig:framework}, we append a projection head $\varphi$ and another projection head $\psi$ to the online decoder and target encoder respectively to produce positive feature pairs $(H_u^Q, Y_v^K)$ and $(H_r^K, Y_w^Q)$. $\varphi$ and $\psi$ both consist of a 3-layer MLPs with $l_2$-normalized bottleneck following DINO~\cite{caron2021emerging}. We then adopt global average pooling to these features and obtain their global visual semantics, denoted as $([cls]_u^Q, [cls]_v^K)$ and $([cls]_r^K, [cls]_w^Q)$.
We encourage their high-level class-specific representations move closer in the corresponding feature space. This yields the loss:
\begin{equation}
  \ell_{dv} = \ell_s([cls]_u^Q, [cls]_v^K) + \ell_s([cls]_r^K, [cls]_w^Q),
  \label{eq:ldv}
\end{equation}
where $\ell_s$ denotes a general cosine similarity loss~\cite{grill2020bootstrap,chen2021exploring}. 
Through optimizing such inter-volume relationship, Alice explicitly enforces anatomical invariance.

\subsection{Anatomical Semantic Alignment}
\label{ASA}
To further borrow the capability of semantics abstraction acquired from self-distillation, we now consider optimizing the intra-volume relationship by maximizing the similarity between views from the same volume. 
Different from recent works~\cite{zhou2021ibot, kakogeorgiou2022hide} 
that use embeddings of intact views ($Y_w^Q$ and $Y_v^K$) as teachers to supervise the masked views’ representations ($V_u^Q$ and $V_r^K$),  
we argue that it is unreasonable to directly encourage the masked views to have similar representations to the global views since their anatomical structure and semantic information may be significantly distinct.
This dilemma motivates us to perform anatomical semantic alignment\footnote{Please note that anatomical semantic alignment does not involve a registration task. Our primary objective is to extract aligned features that demonstrate the highest correlation from masked views and augmented views. We do not aim to match masked images to full images in this process.} between output features from the masked view and the intact view.

One straightforward way is to contrast between the output features of the online decoder and target encoder as proposed in SIM~\cite{tao2022siamese}. However, we empirically observe that this choice brings little improvement for downstream tasks. 
Directly using the features from the online decoder for self-distillation means that the online decoder has to simultaneously optimize multiple different targets. Such multi-task learning process trained a strong online decoder
and thus creates an oversimple optimizing task for the online encoder since the decoder takes the most charge of recovering details and anatomical semantics. 
To ensure that the online encoder effectively contributes to downstream tasks, it is crucial to develop an improved training strategy that enables the online encoder to learn rich and generalized feature representations.
Taking this into account, 
we propose a conditional anatomical semantic alignment (CASA) module with learning capacity of reasoning about the encoded masked view's most semantically similar anatomical features in the embeddings of the original volume. This leads to crafting more specific and aligned high-level features for self-distillation.

We take the process of aligning features from $X_u^Q$ (masked view from volume crop $Q$) and $X_w^Q$ (augmented intact view from volume crop $Q$) as an example to elaborate. 
\cref{fig:framework} shows the high-level idea of the anatomical semantic alignment.
We use $V_u^Q$ from the online encoder as a criterion query, and generate a contrastive pair from outputs of online decoder and target encoder with aligned anatomical semantics constrained by the criterion query. 
Concretely, we generate teacher embeddings $T_w^Q$ and student embeddings $S_u^Q$, which are aligned and capture the most semantically similar anatomical information in the original volume as expressed in the masked view's embeddings $V_u^Q$. Here, the anatomical features from $V_u^Q$ serve as the guidance for semantic alignment.

\begin{figure}
  \centering
  \includegraphics[width=\linewidth]{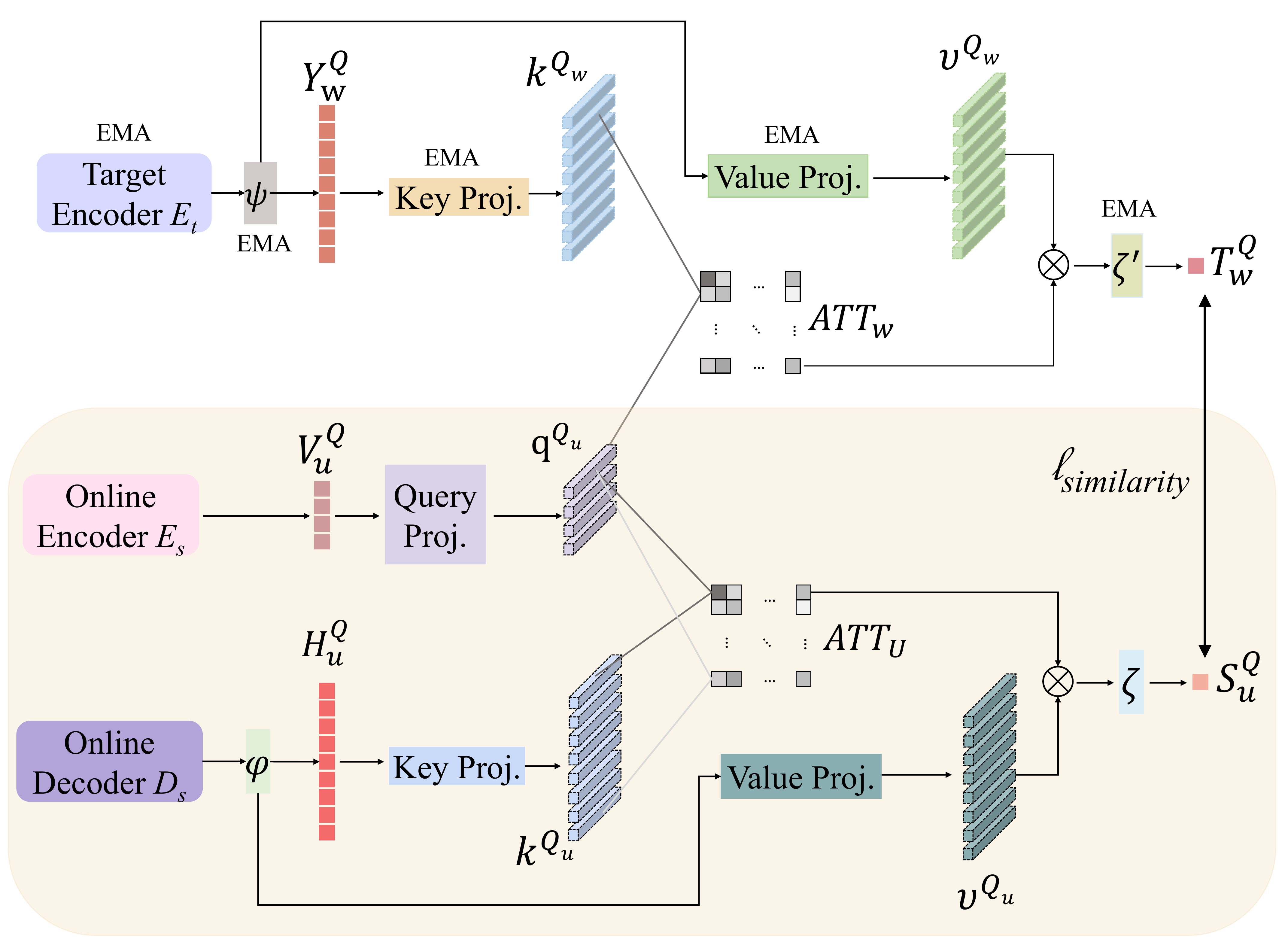}
  \caption{Diagram of the alignment process for volume crop $Q$ in CASA. We generate a semantic-aligned feature pair of student embeddings $S_u^Q$ and teacher embeddings $T_w^Q$. The processes that produce these two embeddings are symmetric.
  We adopt a query projection for $V_u^Q$, key and value projection for both $H_u^Q$ and $Y_w^Q$. We compute the dot product attention between query and key matrices to allow values to match the most relevant high-level semantics given the local patch texture and topology from masked view's feature embeddings ($V_u^Q$).
  }
  \label{fig:CASA}
\end{figure}

\cref{fig:CASA} shows the process of generating the student embeddings $S_u^Q$ (in yellow part) and teacher embeddings $T_w^Q$. Consider generating $S_u^Q$ first,
we project $V_u^Q$ into the query $q^{Q_u}\in \mathbf{R}^{N_m^Q \times C}$ matric: $q^{Q_u} = LN(V_u^Q)W_q$.
Then we project $H_u^Q$, which encodes the reconstructed features, into key $k^{Q_u}\in \mathbf{R}^{N \times C}$ and value $\nu^{Q_u}\in \mathbf{R}^{N \times C}$ matrices: $k^{Q_u} = LN(H_u^Q)W_k$, $\nu^{Q_u} = LN(H_u^Q)W_\nu$.

Here $C$ is the projection dimension, $LN$ is a LayerNorm layer and $W_q$, $W_k$
and $W_\nu$ are projection matrices. 
The query matrices $q^{Q_u}$ are used to seek from the key matrices $k^{Q_u}$ to attend to semantics with the highest relevance. The value matrices $\nu^{Q_u}$ represent the anatomical features from which we aggregate only certain high-level global information depending on $V_u^Q$. Since $V_u^Q$ is extracted from masked views, we aim to adaptively highlights the relevant features from the reconstructed views to selectively obtain anatomical semantics based on their relevance to $V_u^Q$. 

In order to learn flexible conditioning between the online encoder outputs $V_u^Q$ and the reconstructed features $H_u^Q$, we compute scaled dot-product attention~\cite{vaswani2017attention} from the query-projected matrices $q^{Q_u}$ to the key-projected matrices $k^{Q_u}$. The dot product attention gives relevancy weights $ATT_u \in \mathbf{R}^{N_m^Q \times N}$ from local fragmentary patch information to each high-level global anatomical semantics:
\begin{equation}
  ATT_u(V_u^Q, H_u^Q) = softmax \left ( \frac{q^{Q_u} \cdot {k^{Q_u}}^T}{\sqrt{C}} \right ).
  \label{eq:9}
\end{equation}
We then leverage $ATT_u$ to aggregate the value-projected matrices $\nu^{Q_u}$ and get the student embeddings through a projection layer $\zeta$ as follows:
\begin{equation}
  S_u^Q = \zeta(ATT_u \cdot \nu^{Q_u}).
  \label{eq:10}
\end{equation}

The process of generating the teacher embeddings $T_w^Q$ is similar and the mere difference is that we use the target encoder's outputs $Y_w^Q$ to produce key and value matrices. As $T_w^Q$ and $S_u^Q$ are guided by the same query matrices, they learn to encode globally matched anatomical semantics and inter-patch topology information from different views conditioned by the distribution of masked view's local image content.
We then maximize the similarity between semantic-aligned teacher embeddings $T_w^Q$ and student embeddings $S_u^Q$ to produce more suitably learned representations.


The process of aligning features from $X_r^{K}$ (masked view from volume crop $K$) and $X_v^{K}$ (augmented intact view from volume crop $K$) is exactly the same as the above. We also obtain teacher embeddings $T_v^K$ and student embeddings $S_r^K$ from the volume $K$. Now we have positive feature pairs $(S_u^Q, T_w^Q)$ and $(S_r^K, T_v^K)$. 
The semantically aligned features in each pair encompass the same high-level anatomical information, we aim to reach a consensus among their representations by maximizing the similarity between them.
We therefore define the intra-volume loss as:
\begin{equation}
  \ell_{st} = \ell_s(S_u^Q, T_w^Q) + \ell_s(S_r^K, T_v^K),
  \label{eq:lst}
\end{equation}

By involving the inter-volume contrast (in~\cref{AIM}) and intra-volume contrast (in~\cref{ASA}), we explicitly model the anatomical invariance. 
Combining such advantage with the MIM objective, the online encoder learns to capture both high-level discriminative anatomical features and fine-grained localization-sensitive context details.

Note that, in Sec.~\ref{AIM}, we do not adopt semantic alignment between views from different volumes, as they should share the same high-level class information but be distinct in local texture and shape. There is no direct relevance between the masked view from volume crop $Q$ and the intact view from volume crop $K$.
\section{Experiments}
\subsection{Datasets \& Evaluation Metrics}
\noindent \textbf{Pre-training Datasets.} We use a total of 2,000 unlabeled CT scans from the Fast and Low-resource semi-supervised Abdominal oRgan sEgmentation in CT (FLARE 2022) challenge dataset\footnote{https://flare22.grand-challenge.org/} to train {\bf Alice} by self-supervised learning. Any (available) forms of annotations or labels of these 2,000 CTs are \textbf{not} employed during the pre-training stage.

\noindent \textbf{Downstream Datasets \& Evaluation Metrics.}
Three public datasets are used for downstream evaluation. For segmentation task:
(1) In addition to 2,000 unlabeled CT scans, FLARE 2022 also provides a downstream training set including 50 labeled CT scans with pancreatic diseases. 
The offline test set incorporates 20 CT scans of patients with liver, kidney, spleen, or pancreas diseases. The segmentation targets are 13 organs: liver, spleen, pancreas, right kidney, left kidney, stomach, gallbladder, esophagus, aorta, inferior vena cava, right adrenal gland, left adrenal gland, and duodenum. 
(2) The Beyond the Cranial Vault (BTCV) abdomen challenge dataset~\cite{landman2015miccai} contains 30 subjects of abdominal CT scans where 13 organs (not the same as FLARE 2022) are annotated by interpreters under the supervision of radiologists at Vanderbilt University Medical Center. 
Following~\cite{xie2022unimiss}, we employ two settings of test set configurations: offline test set and online test set, for BTCV dataset.
For classification task:
(3) 
We conduct experiments on a public benchmark MosMedData: Chest CT Scans with COVID-19 Related Findings~\cite{morozov2020mosmeddata}. This dataset contains a total of 1,110 lung CT scans with COVID-19 related findings, as well as without such findings. We randomly split $70\%$ of the dataset for training, $10\%$ for validation and the rest $20\%$ for testing.
Note that all downstream datasets do \textbf{not} have any intersection with the dataset used for pre-training.
For quantitative evaluation, we adopt two segmentation metrics of Dice Similarity Coefficient (DSC) and Normalized Surface Dice (NSD), and one classification metric of the area under the receiver operator curve (AUC).

\subsection{Implementation Details}
\noindent \textbf{Pre-training Setup.} We use ViT-B~\cite{dosovitskiy2020image} and Swin-B~\cite{liu2021swin} as the default backbones for online encoder. Other than adopting two different random crops for the online encoder and target encoder as common practice, we utilize SAM~\cite{yan2022sam} to first locate/align the same body part, then we use a default input volume crop size of $192\times192\times64$ to generate respective views of consistent anatomies. This avoids the large disparity between the inputs of online/target encoders when randomly cropped regions are far apart spatially, or scarcely semantically relevant. For MIM tasks, we apply augmentations and masking strategy following MAE~\cite{he2022masked} to the input of the online encoder. For the input of the target encoder, strong data augmentations, \eg random resized rotation, flipping, intensity scaling and shifting, are adopted to avoid a trivial solution. We keep the same projection head structure as used in ~\cite{chen2021empirical, grill2020bootstrap}. The training loss is a summation of $\ell_r$, $\ell_{dv}$, and $\ell_{st}$.
During pre-training, we 
adopt a lightweight decoder to reduce computing overhead and only use the online encoder for downstream segmentation tasks.
We employ AdamW optimizer~\cite{loshchilov2017decoupled} with the momentum set to $\beta_1 = 0.9$, $\beta_2 = 0.95$ and cosine learning rate schedule with a warmup of 100 epochs. The pre-training process uses a batch size of 8 per GPU and an initial learning rate of $5e^{-5}$ for 100K iterations. We implement {\bf Alice} model in PyTorch~\cite{paszke2019pytorch}. All pre-training experiments are conducted on 8 NVIDIA A100 GPUs.

\noindent \textbf{Downstream Training Setup.}
For segmentation tasks, we apply our pre-trained encoder weights to various ViT-based segmentation networks designed for medical tasks of UNETR~\cite{hatamizadeh2022unetr}, nnFormer~\cite{zhou2021nnformer}, and Swin UNETR~\cite{tang2022self},
by following their settings. We compare different SSL methods within ~\cite{hatamizadeh2022unetr,zhou2021nnformer,tang2022self}. Five-fold cross validation is used to train and evaluate models for all FLARE 2022 and BTCV experiments. For classification task, we utilizes $2,000$ unlabeled CT scans from FLARE 2022 for pre-training to evaluate the adaptability of Alice for different scenes (COVID-19 classification). Ten-fold cross validation is used for COVID-19 dataset. Detailed training hyperparameters for these downstream tasks can be found in the supplemental material.

\begin{table*}[htbp]
  \small
  \begin{center}
  \resizebox{0.9\linewidth}{!}
  {
  \begin{tabular}{@{}l|cc|cc|cc@{}}
    \toprule[1pt]
    \multicolumn{1}{l}{\multirow{2}{*}{\diagbox{Method}{Backbone}}} & \multicolumn{2}{|c}{UNETR} & \multicolumn{2}{c}{Swin UNETR} & \multicolumn{2}{c}{nnFormer}\\
    \cmidrule{2-3}\cmidrule{4-5}\cmidrule{6-7}           
    & DSC & NSD & DSC & NSD & DSC & NSD \\
    \midrule
    Rand. init.                              & 80.95$\pm$3.48    & 85.23$\pm$3.74     & 81.04$\pm$3.29   & 85.60$\pm$3.55   & 81.33$\pm$3.05  &  86.05$\pm$3.31\\
    MoCo v3~\cite{chen2021empirical}         & 82.01$\pm$3.27    & 86.49$\pm$3.52     & 82.63$\pm$3.18   &  86.92$\pm$3.44   & 82.88$\pm$2.82  &  86.89$\pm$3.10\\
    DINO~\cite{caron2021emerging}            & 82.07$\pm$3.15    & 86.31$\pm$3.40     & 82.69$\pm$2.99   &  87.08$\pm$3.17   & 82.95$\pm$2.74  &  87.45$\pm$2.92\\
    IBOT~\cite{zhou2021ibot}                 & 83.15$\pm$3.21    & 87.61$\pm$3.48     & 83.77$\pm$3.14   &  88.34$\pm$3.30   & 84.04$\pm$2.81  &  88.46$\pm$3.11\\
    SIM~\cite{tao2022siamese}                & 83.04$\pm$2.97    & 87.37$\pm$3.36     & 83.59$\pm$2.74   &  88.57$\pm$2.90   & 83.96$\pm$2.64  &  88.61$\pm$2.83 \\ 
    MAE~\cite{he2022masked}                  & 83.09$\pm$2.92    & 87.42$\pm$3.43     & 83.62$\pm$2.66   &  88.55$\pm$2.92   & 84.01$\pm$2.59  &  88.63$\pm$2.75\\
    SemMAE~\cite{li2022semmae}               & 83.13$\pm$2.87    & 87.90$\pm$3.31     & - & - & - & - \\
    CMAE~\cite{huang2022contrastive}         & 83.59$\pm$2.76    & 88.25$\pm$2.98     & 84.25$\pm$2.59   & 89.17$\pm$2.88   & 84.47$\pm$2.42  &  89.44$\pm$2.65\\
    \midrule
    medical MAE~\cite{zhou2022self}           & 83.11$\pm$2.91    & 87.45$\pm$3.39   &83.66$\pm$2.64    & 88.57$\pm$2.86    & 84.03$\pm$2.57  & 88.67$\pm$2.71 \\
    Tang \textit{et al.}~\cite{tang2022self}  & 82.97$\pm$3.22    & 87.51$\pm$3.50     & 83.14$\pm$3.01   & 88.74$\pm$3.37   & 83.59$\pm$2.89  & 89.51$\pm$3.23\\
    \midrule
    {\bf Alice}         & \textbf{85.81}$\pm$\textbf{2.05}   & \textbf{90.03}$\pm$\textbf{2.28}  & \textbf{86.75}$\pm$\textbf{1.89} & \textbf{91.22}$\pm$\textbf{2.12} & \textbf{86.87}$\pm$\textbf{1.84} & \textbf{91.28}$\pm$\textbf{2.09} \\
    \bottomrule
  \end{tabular}
  }
  \end{center}
  \caption{Average DSC and NSD of 13 organs obtained using different ViT-based SSL strategies on the FLARE 2022 offline test set. We fix the adopted segmentation baselines as UNETR, nnFormer, and Swin UNETR. ``-'' means there is no direct adaptation on the corresponding Swin Transformer based backbones.}
  \label{tab:flare1}
\end{table*}

\subsection{Results}

\noindent \textbf{FLARE 2022 Abdominal Organ Segmentation.}
We conduct extensive comparison between {\bf Alice} and existing SOTA methods. We first fix three ViT-based medical segmentation framework, and compare {\bf Alice} with the random initialization (Rand.~init.) and other advanced SSL methods designed for computer vision and medical tasks. The evaluation results on the offline test set are shown in Table~\ref{tab:flare1}.
Compared with strong Transformer-based contrastive learning methods MoCo v3 and DINO, 
{\bf Alice} outperforms them by at least absolute $3.74\%$ in DSC. Compared with MIM methods MAE and SemMAE, the DSC score of Alice surpasses that of MAE and SemMAE by a large margin. Furthermore, {\bf Alice} also achieves at least $2.22\%$ improvement of DSC relative to hybrid SSL methods IBOT and CMAE, which combine contrastive learning and MIM. Notably, Alice is also superior to SOTA ViT based SSL methods, medical MAE~\cite{zhou2022self} and pre-training framework proposed in~\cite{tang2022self}, which are tailored for medical image analysis. 

\begin{table}[htbp]
  \Huge
  \begin{center}
  \resizebox{\linewidth}{!}
  {
  \begin{tabular}{@{}lccc@{}}
    \toprule[2pt]
    Method & Pre-trained Parts & DSC  & NSD\\
    \midrule
    MoCo v2~\cite{chen2020improved} & 3D ResNet E.                & 81.96$\pm$3.29 & 86.58$\pm$3.45\\
    BYOL~\cite{grill2020bootstrap} & 3D ResNet E.                 & 81.94$\pm$3.32 & 86.60$\pm$3.50\\
    ContrastiveCrop~\cite{peng2022crafting} & 3D ResNet E.        & 82.55$\pm$3.05 & 87.26$\pm$3.15\\
    LoGo~\cite{zhang2022leverage} & 3D ResNet E.                  & 82.53$\pm$3.08 & 87.08$\pm$3.17\\
    \midrule
    PCRL~\cite{zhou2021preservational} & 3D UNet E.\&D.           & 83.41$\pm$2.81 & 87.84$\pm$3.00\\
    PCRLv2~\cite{zhou2023unified} & 3D nsUNet E.                & 84.45$\pm$2.49  & 89.11$\pm$2.68 \\
    PGL~\cite{xie2020pgl} & 3D ResNet E.                          & 83.16$\pm$2.85 & 87.93$\pm$2.99\\
    Chaitanya~\etal~\cite{chaitanya2020contrastive} & UNet E.\&D. & 82.29$\pm$2.90 & 86.72$\pm$3.14 \\
    nnU-Net~\cite{isensee2021nnu} & -                             & 83.62$\pm$2.78 & 88.51$\pm$2.85\\
    TransVW~\cite{haghighi2021transferable} & 3D UNet E.\&D.      & 82.70$\pm$3.02 & 87.26$\pm$3.18\\
    SAM~\cite{yan2022sam}           & 3D UNet E.\&D.              & 82.48$\pm$3.06 & 87.06$\pm$3.21\\
    Models Gen.~\cite{zhou2021models} & V-Net E.\&D.              & 83.35$\pm$2.82 & 88.12$\pm$2.95\\
    DiRA~\cite{haghighi2022dira} & 3D UNet E.\&D.                 & 82.57$\pm$2.88 & 87.23$\pm$3.01  \\  
    \midrule
    {\bf Alice} & nnFormer E.  & \textbf{86.87}$\pm$\textbf{1.84} &  \textbf{91.28}$\pm$\textbf{2.09}\\
    \bottomrule[2pt]
  \end{tabular}
  }
  \end{center}
  \caption{Average DSC and NSD of 13 organs obtained using different CNN-based SSL strategies and models on the FLARE 2022 offline test set. ``E.'' means only encoder is pre-trained while ``E.\&D.'' means both encoder and decoder are pre-trained.}
  \label{tab:flare2}
\end{table}

\begin{table}[htbp]
\Huge
  \begin{center}
  \resizebox{\linewidth}{!}
  {
  \begin{tabular}{@{}lccc@{}}
    \toprule[2pt]
    Method  & Ensemble & Offline Test Set & Online Test Set\\
    \midrule
    MoCo v2~\cite{chen2020improved}           & 1       & 82.05$\pm$2.82          & - \\
    MoCo v3~\cite{chen2021empirical}          & 1       & 82.02$\pm$2.77          & - \\
    DINO~\cite{caron2021emerging}             & 1       & 82.61$\pm$1.79          & - \\
    MAE~\cite{he2022masked}                   & 1       & 83.16$\pm$2.14          & - \\
    IBOT~\cite{zhou2021ibot}                  & 1       & 83.28$\pm$2.26          & - \\
    CMAE~\cite{huang2022contrastive}          & 1       & 83.47$\pm$1.33          & - \\
    \midrule
    PCRL~\cite{zhou2021preservational}        & 1       & 82.73$\pm$2.42          & - \\
    PCRLv2~\cite{zhou2023unified}             & 1       & 83.55$\pm$1.49          & - \\
    PGL~\cite{xie2020pgl}                     & 1       & 82.57$\pm$2.60          & - \\
    Chaitanya~\etal~\cite{chaitanya2020contrastive} & 1 & 82.74$\pm$2.12 & - \\
    medical MAE~\cite{zhou2022self}           & 1       & 83.23$\pm$2.05          & - \\
    SAM~\cite{yan2022sam}                     & 1       & 82.00$\pm$3.01          & 84.07 \\
    DoDnet~\cite{zhang2021dodnet}             & 5       & -         & 86.44 \\
    UNETR~\cite{hatamizadeh2022unetr}         & 5       & -         & 85.55 \\
    PaNN~\cite{zhou2019prior}                 & 5       & -         & 85.00 \\
    nnU-Net~\cite{isensee2021nnu}             & 10      & -         & 87.62    \\
    nnFormer~\cite{zhou2021nnformer}         & 5        & 82.88$\pm$2.59         & -  \\
    DiRA~\cite{haghighi2022dira}              & 1       & 83.14$\pm$2.04    & -     \\
    Tang \etal~\cite{tang2022self}             & 1       & 82.58$\pm$2.20   & 84.72     \\
    UniMiSS~\cite{xie2022unimiss}             & 1       & 84.99$\pm$1.57         & 87.05      \\
    \midrule
    {\bf Alice}  & 1 & \textbf{86.76}$\pm$\textbf{0.98} &  \textbf{88.58}\\
    \bottomrule[2pt]
  \end{tabular}
  }
  \end{center}
  \caption{Average DSC of 13 organs on the BTCV offline and online test set. Most results of online test set are directly obtained from BTCV test leaderboard. Result $84.72$ of Tang~\etal is drawn from their paper. Results of MoCo v2, MoCo v3, DINO, PCRL, PGL and UniMiSS are drawn from the original paper of UniMiSS. The segmentation backbone for {\bf Alice} is nnFormer.}
  \label{tab:BTCV1}
\end{table}

\begin{figure*}
  \centering
  \includegraphics[width=\linewidth]{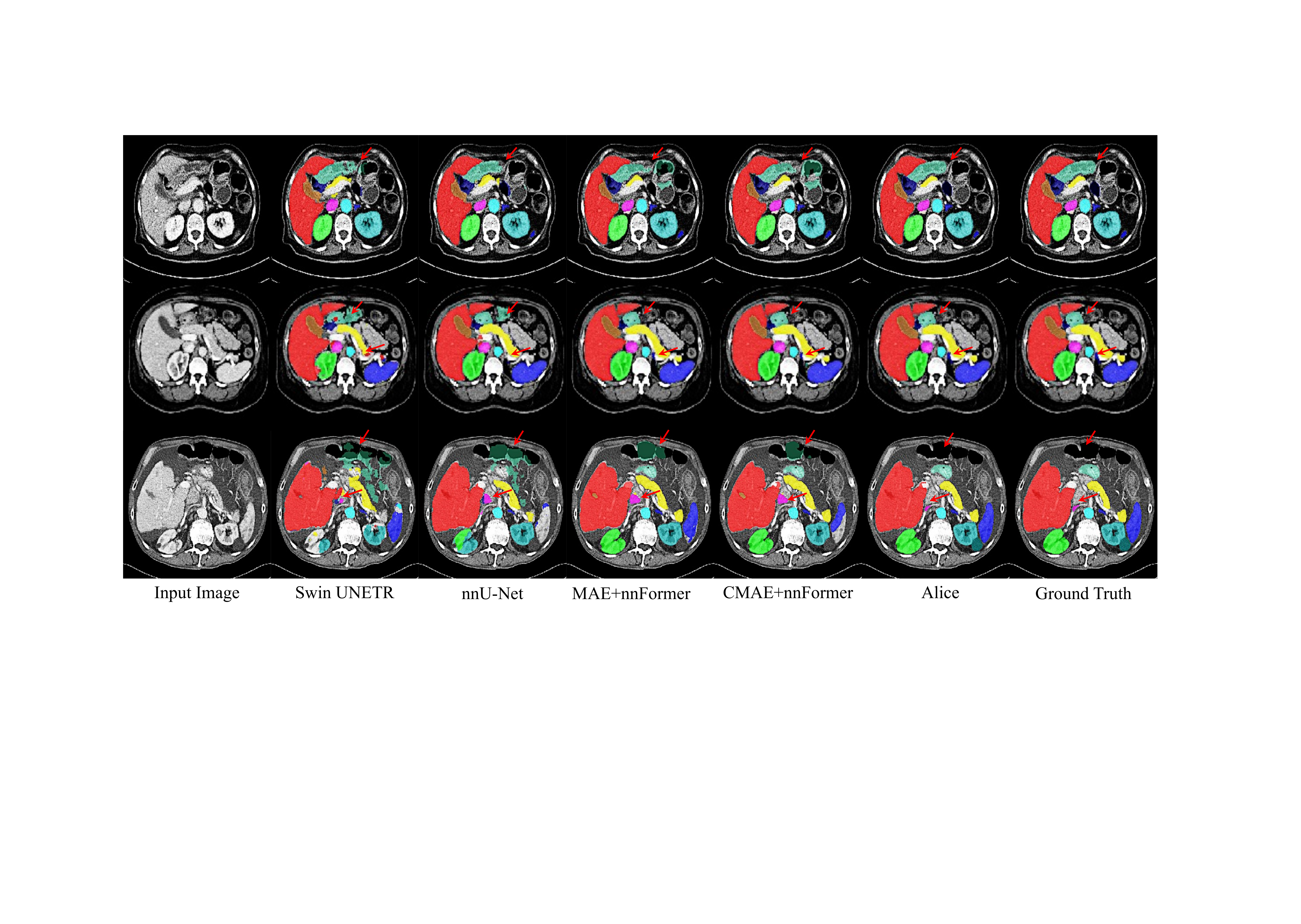}
  \caption{Qualitative visualizations on FLARE 2022 offline test set. We compare Alice with other advanced SSL methods MAE, CMAE, self-supervised Swin UNETR, and strong CNN baseline nnU-Net. 
  We present the visualizations of BTCV in the supplemental material.}
  \label{fig:flare22results}
\end{figure*}


\begin{table}[htbp]
\Large
  \centering
  \resizebox{\linewidth}{!}
  {
  \begin{tabular}{@{}lc|lc@{}}
    \toprule[1.5pt]
    CNN-based Method & AUC on COVID-19 & ViT-based Method & AUC on COVID-19\\
    \hline
    BYOL~\cite{grill2020bootstrap}                 & 85.74$\pm$5.04     & DINO~\cite{caron2021emerging}        & 86.87$\pm$4.35  \\
    Peng~\etal~\cite{peng2022crafting}             & 87.02$\pm$3.11      & IBOT~\cite{zhou2021ibot}           & 87.55$\pm$3.63   \\
    LoGo~\cite{zhang2022leverage}                  & 86.95$\pm$3.59     & SIM~\cite{tao2022siamese}              & 87.67$\pm$2.95   \\
    PCRL~\cite{zhou2021preservational}             & 87.31$\pm$2.88     & MAE~\cite{he2022masked}                & 86.62$\pm$3.27\\
    PCRLv2~\cite{zhou2023unified}                  & 88.36$\pm$2.51       & SemMAE~\cite{li2022semmae}               & 86.94$\pm$3.44    \\
    PGL~\cite{xie2020pgl}                          & 86.08$\pm$4.72     & CMAE~\cite{huang2022contrastive}       & 87.73$\pm$3.02   \\
    DiRA~\cite{haghighi2022dira}                   & 87.43$\pm$3.55      & {\bf Alice}            & \textbf{90.88}$\pm$\textbf{1.29}  \\
    \bottomrule[1.5pt]
  \end{tabular}
  }
  \caption{Classification performance of using different pre-training strategies on 
  the COVID-19 screening test set. CNN-based SSL methods take the 3D ResNet as their encoder backbone.
  ViT-based SSL methods take the ViT-B as their encoder backbone.}
  \label{tab:COVID19}
\end{table}

We also compare Alice with strong CNN-based SSL methods and baselines. The results are shown in Table~\ref{tab:flare2}. Alice significantly outperforms the other CNN-based SOTA methods. {\bf Alice} achieves better performance compared with PCRL, TransVW, SAM, DiRA, and Models genesis, which additionally pre-train a decoder. Notably, {\bf Alice} outperforms the method of Chaitanya~\etal~\cite{chaitanya2020contrastive} which also utilizes inter-volume slices. Chaitanya~\etal~\cite{chaitanya2020contrastive} assume that all
volume images have been aligned and use simple grouping of image slices in 3D volumes to craft contrastive patches, which will lead to unaligned pairs. Differently, we adopt a learning-based module SAM~\cite{yan2022sam} to get aligned sample pairs from different volumes online and we further align feature level semantics by CASA module. Our Alice is more suitable to handle real-world complex problems like changes in the range of CT scans and temporal changes.


\cref{fig:flare22results} shows the qualitative results and demonstrate the merits of {\bf Alice}. Most competing methods suffer from segmentation target incompleteness related failures and misclassification of background regions as organs (false positives). {\bf Alice} produces sharper boundaries and generates results that are more consistent with the ground truth in comparison with all other models. 
This success is attributed to the advantage of leveraging the intrinsic anatomical invariance across varying volumes and semantic alignment between contrastive views, which help the learned representation robust to organ deformation and pathological changes. 


\noindent \textbf{BTCV Multi-organ Segmentation.}
We also compare {\bf Alice} with other SOTA SSL methods on the BTCV offline and online test set. As shown in Table~\ref{tab:BTCV1}, {\bf Alice}, without using any ensemble strategy, still achieves the competitive performance with the best DSC on both offline and online test set.
Note that compared to SOTA methods designed for medical images, namely UniMiSS and the self-supervised Swin UNETR (Tang \textit{et al.}), using over $5,000$ 3D CT scans for pre-training, our {\bf Alice} outperforms them using only $40\%$ of the data. This effectiveness can be explained that our approach is capable of learning representations that is robust to the size, shape, intensity, and texture diversity of body parts by modeling the anatomical invariance. During pre-training, we map the high-level embeddings of the same organs from varying volumes to the same point, which helps the model in downstream tasks reduce misclassification failures and noticeably improves the performance.

\noindent \textbf{COVID-19 Classification.}
We compare {\bf Alice} with the state-of-the-arts including representative CNN-based SSL methods and ViT-based SSL methods. The results are shown in~\cref{tab:COVID19}. 
Compared with CNN-based SSL methods BYOL, PCRL, PCRLv2, PGL and DiRA, {\bf Alice} outperforms them at least absolute $2.52\%$ in AUC. Notably, {\bf Alice} achieves much better results than LoGo~\cite{zhang2022leverage} and Peng~\etal~\cite{peng2022crafting}, which also design specific strategies to generate semantic-aligned contrastive view pairs. However, these two methods only operate within each image independently and ignore the inter-volume consistency. Compared against strong ViT-based SSL methods, {\bf Alice} significantly outperforms them at least absolute $3.15\%$ in AUC. It proves the effectiveness of modeling anatomical invariance and performing semantic alignment to assist the SSL process. Moreover, as an inter-scene evaluation, \cref{tab:COVID19} reveals {\bf Alice} has the ability to generalize to other scenarios. More details are in our Supplementary Material.

\subsection{Ablation Study}
\noindent \textbf{Significance in Modeling Inter-volume Anatomical Invariance.}
We investigate the effect of leveraging the intrinsic anatomical structures of different volumes to model the anatomical invariance. We feed the same body part of query volume and key volume to another SSL method IBOT and add the inter-volume relationship contrastive loss $\ell_{dv}$ for it. Table~\ref{tab:cpm} shows that involving feature contrasting between mined consistent views from varying aligned CT volumes can substantially improve the 3D segmentation accuracy in the downstream task for both IBOT and {\bf Alice}. The performance gain on DSC is $1.37\%$ and $1.24\%$, respectively. This shows that modeling the anatomical invariance among CT volumes to learn intrinsic high-level semantics during pre-training benefits the learned representations for downstream tasks.

\begin{table}
\small
  \begin{center}
  \resizebox{0.85\linewidth}{!}
  {
  \begin{tabular}{@{}lccc@{}}
    \toprule[0.8pt]
    \multirow{2}{*}{Method}  & \multirow{2}{*}{\makecell[c]{Inter-Volume\\Relationship $\ell_{dv}$}}  & \multicolumn{2}{c}{FLARE 2022} \\ 
    \cmidrule{3-4}
    & & DSC & NSD \\
    \midrule
    \multirow{2}{*}{IBOT~\cite{zhou2021ibot}}           & $\times$      &  84.04$\pm$2.81   & 88.46$\pm$3.11    \\
                                    & \checkmark    &  85.41$\pm$2.42    & 89.97$\pm$2.70             \\
    \multirow{2}{*}{{\bf Alice}}          & $\times$      & 85.63$\pm$2.11    & 90.01$\pm$2.32              \\
                                    & \checkmark    &  \textbf{86.87}$\pm$\textbf{1.84} & \textbf{91.28}$\pm$\textbf{2.09}              \\
    \bottomrule[0.8pt]
  \end{tabular}
  }
  \end{center}
  \caption{Ablation study of modeling the inter-volume anatomical invariance. The segmentation backbone is nnFormer. ``$\times$'' means inputing views from the same volume and not using $\ell_{dv}$, while ``\checkmark'' means inputing views from different volumes and using $\ell_{dv}$.}
  \label{tab:cpm}
\end{table}

\noindent \textbf{Effectiveness of Anatomical Semantic Alignment.}
Table~\ref{tab:casa} shows the ablation study on the effectiveness of the CASA module. Our approach can be applied to most general siamese architecture based SSL methods. All methods achieve more than $1.21\%$ absolute DSC gain on the FLARE 2022 dataset by adopting CASA. This shows applying our anatomical semantic alignment to harvest better contrastive pairs from masked and unmasked views yields significant improvements on various siamese SSL methods.

\begin{table}
\Huge
  \begin{center}
  \resizebox{\linewidth}{!}
  {
  \begin{tabular}{@{}lccc@{}}
    \toprule[2.5pt]
    \multirow{2}{*}{Method}  & \multicolumn{2}{c}{FLARE 2022}  \\ 
    \cmidrule{2-3}
     & DSC & NSD  \\
    \midrule
    IBOT (\textit{w}\textit{/}.~vs.~\textit{w}\textit{/}\textit{o}.) & 85.34$\pm$2.63~vs.~84.04$\pm$2.81 &89.88$\pm$2.89~vs.~88.46$\pm$3.11  \\
    SIM (\textit{w}\textit{/}.~vs.~\textit{w}\textit{/}\textit{o}.)  & 85.21$\pm$2.36~vs.~83.96$\pm$2.64  & 89.89$\pm$2.67~vs.~88.61$\pm$2.83 \\
    \midrule
    {\bf Alice} (\textit{w}\textit{/}.~vs.~\textit{w}\textit{/}\textit{o}.)   & \textbf{86.87}$\pm$\textbf{1.84}~vs.~85.66$\pm$2.18  &  \textbf{91.28}$\pm$\textbf{2.09}~vs.~90.02$\pm$2.32   \\
    \bottomrule[2.5pt]
  \end{tabular}
  }
  \end{center}
  \caption{Ablation study of the CASA on FLARE 2022 benchmark. 
    We apply the CASA on different siamese SSL architectures which take masked view and unmasked view as inputs. ``\textit{w}\textit{/}.~vs.~\textit{w}\textit{/}\textit{o}.'' denotes the comparison
    between using CASA or without using CASA. The segmentation backbone for {\bf Alice} is nnFormer.}
  \label{tab:casa}
\end{table}

\noindent \textbf{The impact of whether to use negative samples.}
In Eq.~\ref{eq:ldv}, we utilize BYOL-style cosine loss~\cite{grill2020bootstrap} as our default choice for contrastive learning. This loss only maximizes the similarity between positive views and eliminates the use of negative pairs. Another widely used similarity loss function is the InfoNCE loss~\cite{chen2020simple,he2020momentum, oord2018representation}, which aims to simultaneously pull close positive views and push away negative samples. The key distinction between these two widely used types of similarity loss functions lies in the utilization of negative samples. {\bf Alice} is compatible with a wide range of SSL techniques and is independent of the specific training losses used in those techniques. Thus we conduct experiments to investigate the influence of whether negative pairs are used. We discuss two widely used similarity loss functions: InfoNCE loss (exploits both positive and negative samples) and BYOL-style cosine loss (does not exploit negative samples). When using InfoNCE loss for contrastive learning, we utilize views from crops of different body parts in the same batch to compose negative pairs. 
Table~\ref{tab:ls} shows the ablation study on whether to use negative samples. We observed that using cosine loss in {\bf Alice} achieves slightly higher performance than InfoNCE loss on FLARE 2022 test set. Thus, we do not utilize negative samples and use cosine loss as the default for the contrastive learning branch in {\bf Alice}. 
The discussion of why cosine loss does not need negative pairs can be found in BYOL’s literature~\cite{grill2020bootstrap}.

\begin{table}
  \begin{center}
  \resizebox{\linewidth}{!}
  {
  \begin{tabular}{@{}lccc@{}}
    \toprule[1pt]
    \multirow{2}{*}{Method}  & \multirow{2}{*}{loss function $\ell_s$} & \multicolumn{2}{c}{FLARE 2022} \\ 
    \cmidrule{3-4}
    & & DSC & NSD \\
    \midrule
    \multirow{2}{*}{{\bf Alice}}          & InfoNCE loss~\cite{chen2020simple,he2020momentum, oord2018representation}      & 86.83$\pm$1.88    & 91.20$\pm$2.12              \\
                                    & BYOL-style cosine loss~\cite{grill2020bootstrap}    &  \textbf{86.87}$\pm$\textbf{1.84} & \textbf{91.28}$\pm$\textbf{2.09}              \\
    \bottomrule[1pt]
  \end{tabular}
  }
  \end{center}
  \caption{Ablation study on whether to use negative samples. The segmentation backbone is nnFormer. InfoNCE loss seeks to simultaneously pull close positive views and push away negative samples while BYOL-style cosine loss only maximizes the similarity between positive views and eliminates the use of negative pairs.}
  \label{tab:ls}
\end{table}

\label{sec:experiment}
\section{Conclusions}
In this work, we propose a novel self-supervised learning method ({\bf Alice}) for improving the learned image representation of contrastive learning and MIM by modeling the class-specific invariance of intrinsic anatomical semantics in 3D medical images. We also introduce a conditional anatomical semantic alignment module that generates better contrastive pairs with more consistent high-level information. Extensive quantitative experiments reporting superior results validate the effectiveness of our method.
\label{sec:conclusion}




{\small
\bibliographystyle{ieee_fullname}
\bibliography{egpaper_for_review}
}

\end{document}